\documentclass[letterpaper, 10 pt, conference]{ieeeconf}  

\IEEEoverridecommandlockouts                              

\usepackage{graphics} 
\usepackage{epsfig} 
\usepackage{mathptmx} 
\usepackage{times} 
\usepackage{amsmath} 
\usepackage{amssymb}  
\usepackage{multirow}  

\usepackage{cite}
\usepackage{amsmath}

\usepackage{booktabs}
\usepackage{makecell}
\usepackage{graphicx}

\usepackage{multirow}
\usepackage{colortbl}
\usepackage{xcolor}

\title{\LARGE \bf
TempoFit: Plug-and-Play Layer-Wise Temporal KV Memory for Long-Horizon Vision-Language-Action Manipulation}

\author{
\begin{tabular}{c}
Jun Sun$^{1*}$ \quad
Boyu Yang$^{1*}$ \quad
Jiahao Zhang$^{1}$ \quad
Ning Ma$^{1}$ \quad
Chencheng Wu$^{1}$ \\[0.3em]
Siqing Zhang$^{1}$ \quad
Yiou Huang$^{1}$ \quad
Qiufeng Wang$^{1}$ \quad
Shan Liang$^{1}$ \quad
Yaran Chen$^{1}$ \\[0.6em]
$^{1}$Xi'an Jiaotong-Liverpool University \\[0.2em]
$^{*}$Equal contribution
\end{tabular}
}

\newcommand{\FGTB}{\textsc{FGTB}}

\begin{document}

\bstctlcite{BSTcontrol}

\maketitle
\thispagestyle{empty}
\pagestyle{empty}

\begin{figure*}[t]
\vspace{2mm}
  \centering
  \includegraphics[width=\textwidth]{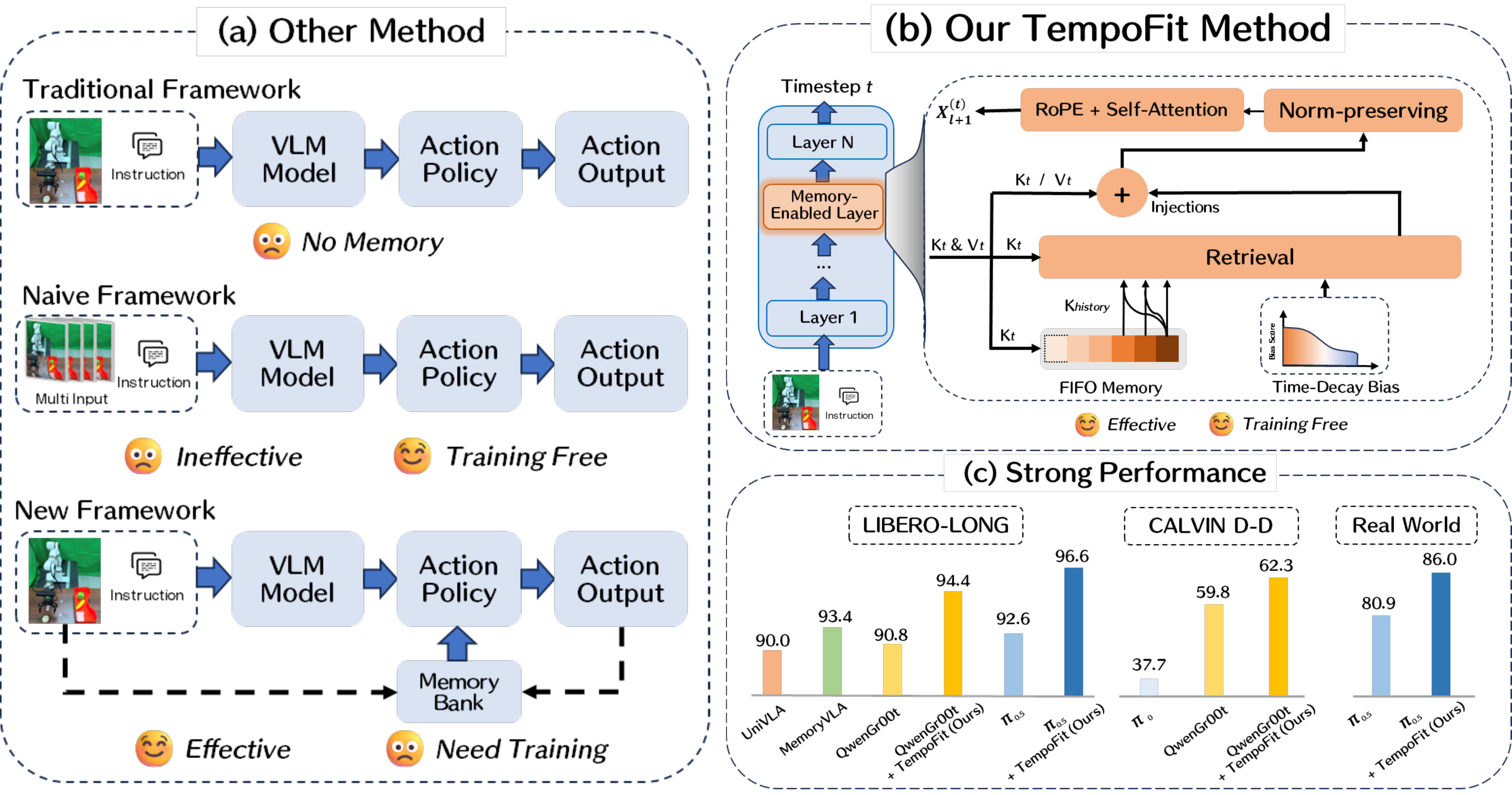}
  \caption{\textbf{TempoFit overview.} At each timestep, TempoFit caches prefix $K/V$ at selected intermediate layers, retrieves relevant history via K-to-K matching with \FGTB{}, and injects the retrieved context through pre-attention residual loading (optionally with norm-preserving rescaling), enabling training-free temporal retrofitting without expanding input context length.}
  \label{fig:overview}
\end{figure*}

\begin{abstract}
Pretrained Vision--Language--Action (VLA) policies have achieved strong single-step manipulation, but their inference remains largely \emph{memoryless}, which is brittle in non-Markovian long-horizon settings with occlusion, state aliasing, and subtle post-action changes. Prior approaches inject history either by stacking frames, which scales visual tokens and latency while adding near-duplicate pixels, or by learning additional temporal interfaces that require (re-)training and may break the original single-frame inference graph.
We present \textbf{TempoFit}, a \emph{training-free} temporal retrofit that upgrades frozen VLAs through \emph{state-level} memory. Our key insight is that prefix attention $K/V$ already form a model-native, content-addressable runtime state; reusing them across timesteps introduces history without new tokens or trainable modules. TempoFit stores layer-wise FIFO prefix $K/V$ at selected intermediate layers, performs parameter-free K-to-K retrieval with Frame-Gap Temporal Bias (FGTB), a fixed recency bias inspired by positional biases in NLP, to keep decisions present-dominant, and injects the retrieved context via pre-attention residual loading with norm-preserving rescaling to avoid distribution shift under frozen weights.
On \textsc{LIBERO-Long}, TempoFit improves strong pretrained backbones by up to \textbf{+4.0}\% average success rate while maintaining near-real-time latency, and it transfers consistently to \textsc{CALVIN} and real-robot long-horizon tasks.
Code is available at \texttt{https://github.com/LucioSunj/TempoFit}.
\end{abstract}

\section{INTRODUCTION}

In recent years, Vision--Language--Action (VLA) models~\cite{zitkovich2023rt2,kim2024openvla,black2024pi0,li2024cogact,wen2025dexvla}
have emerged as a promising framework for robotic manipulation by leveraging large pretrained vision--language backbones~\cite{bai2023qwen,beyer2024paligemma}
to map visual--linguistic representations to the action space.

Despite rapid progress, most mainstream VLA models still perform inference in a largely \emph{memoryless} manner, effectively following a \emph{single-frame decision} paradigm: at each step they encode only the current observation and instruction and directly predict the next action.
This implicitly assumes a Markovian setting, whereas real robot operations are often partially observable and non-Markovian, where the current frame alone may be insufficient to determine the correct action.
In scenarios with occlusion, state aliasing, or when visual changes after actions are subtle, models are prone to failure modes such as repeated operations, missed steps, and cross-stage discontinuity.

Recent work~\cite{octo2024,cheang2024gr2} attempts to mitigate temporal short-sightedness by expanding the observation context, most commonly via stacking a short history of frames.
However, this \emph{observation-level} temporal modeling is structurally inefficient for VLAs: it increases the number of visual tokens and thus the attention compute footprint, leading to higher inference latency, while much of the added signal is near-duplicate pixels that introduce substantial redundancy and can obscure task-relevant dynamics~\cite{lin2025hifvla,koo2025hamlet}.

Beyond stacking, many approaches avoid raw-frame redundancy by encoding history into compact representations and injecting them through retrieval-and-fusion~\cite{Jang2025ContextVLA,shi2025memoryvla,koo2025hamlet}.
However, this shifts temporal modeling to a new learned interface that is not part of the original single-frame inference graph; without training or fine-tuning, the backbone and action head have no guarantee to interpret this new state correctly.
As a result, these methods are generally not directly plug-and-play for strong pretrained single-frame VLAs when weights are frozen, limiting scalable deployment.

These trade-offs leave a clear gap: we still lack a temporal enhancement that can \emph{retrofit} strong pretrained VLAs with history awareness \emph{without} expanding the input context, introducing trainable modules, or requiring additional training.
To fill this gap, we propose \textbf{Layer-Wise Temporal KV Memory}, a \textbf{training-free} inference-time module that injects temporal consistency by reusing the backbone’s \emph{internal} attention state.
Our key idea is to treat the prefix attention \emph{keys/values} ($K/V$) produced during vision--language encoding as a compact, model-native carrier of past context, analogous to the KV-cache-centric view of long-context language-model inference.
Rather than storing raw frames or learning an external memory interface, we cache and reuse prefix $K/V$ only at a \emph{selected subset of intermediate layers}, balancing temporal continuity with minimal interference to present-step control.
At each step, we retrieve relevant historical $K/V$ via lightweight similarity matching and \emph{residually load} the retrieved context into the current step’s $K/V$ \emph{before} standard self-attention (Fig.~\ref{fig:overview}), preserving the pretrained parameters and tokenization.
To keep the current observation \emph{present-dominant} in a training-free setting, we propose \textbf{Frame-Gap Temporal Bias (\FGTB{})}, a fixed frame-gap recency bias inspired by positional biases in NLP~\cite{press2022alibi}, which imposes an explicit decay on retrieval scores without learned gates.


We evaluate TempoFit extensively on long-horizon manipulation benchmarks and real-world robotic tasks, demonstrating consistent gains without additional training.
On \textsc{LIBERO-Long}, TempoFit improves a strong pretrained $\pi_{0.5}$ baseline from 92.6\% to 96.6\% (\textbf{+4.0} abs.) and also boosts a heterogeneous \textsc{StarVLA} checkpoint (QwenGR00T) from 90.8\% to 94.4\% (\textbf{+3.6} abs.), reaching performance competitive with representative training-based temporal models\cite{shi2025memoryvla,lin2025hifvla} while remaining plug-and-play.
On \textsc{CALVIN}, TempoFit improves long-horizon sequential execution in both settings, increasing average task length from 3.78 to 3.84 on D-D and from 3.83 to 3.87 on ABC-D, with clearer gains on later instructions.
Finally, we show that the proposed temporal retrieval and fusion introduces only a negligible inference-time overhead, preserving real-time control (Table.~\ref{tab:efficiency}).

Our main contributions are summarized as follows:
\begin{itemize}
   \item We propose \textbf{TempoFit}, a \emph{training-free} inference-time temporal retrofit that improves temporal consistency and long-horizon manipulation in pretrained VLA policies \emph{without} changing model parameters, training objectives, or input context length.

   \item We introduce a \textbf{layer-wise} KV-native retrieval-and-injection operator with \FGTB{}, a fixed frame-gap recency bias, which suppresses stale context and reduces history--present interference under frozen weights.

   \item Extensive experiments show that our approach improves long-horizon success on widely used benchmarks while preserving high inference efficiency under real-time constraints.
\end{itemize}

\begin{figure*}[t]
\vspace{3mm}
  \centering
      \includegraphics[width=0.98\textwidth]{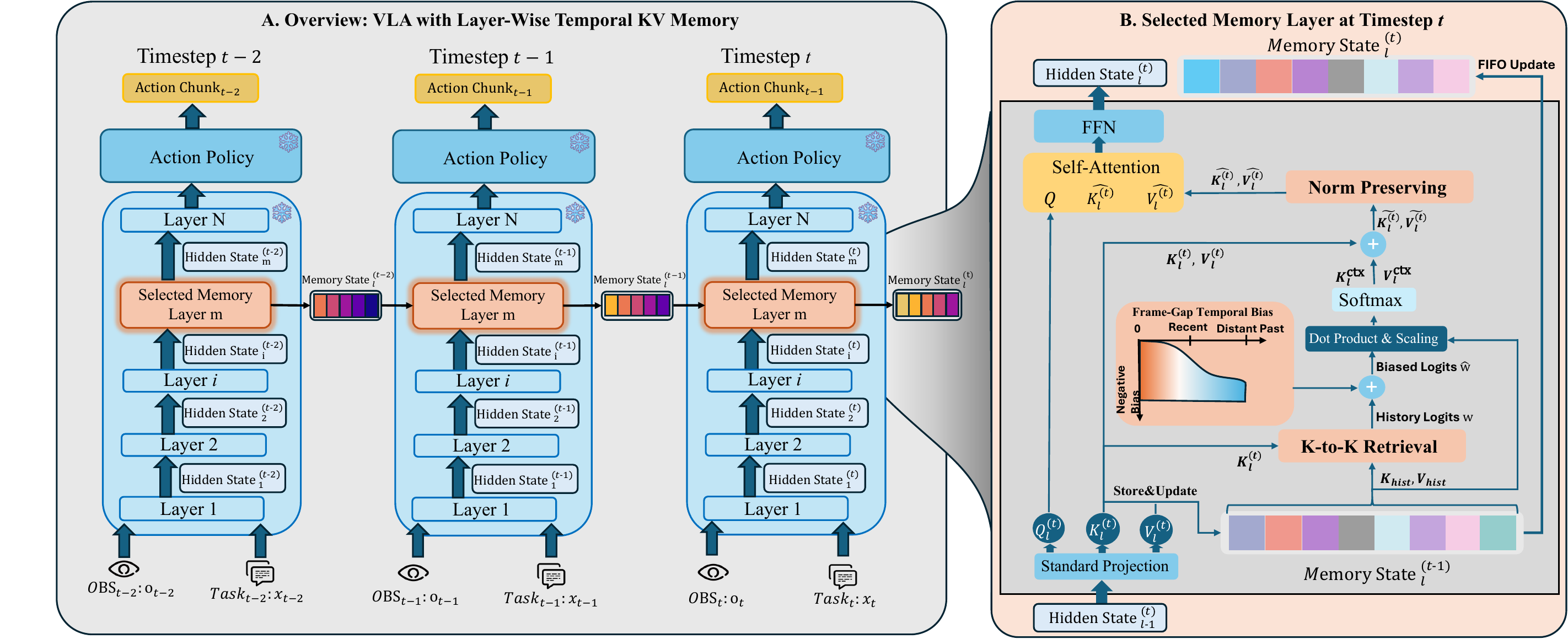}
  \caption{\textbf{TempoFit Pipeline.} (a) In Layer-Wise FIFO KV Cache (see Sec. \ref{sec:write}), TempoFit caches prefix $K/V$ states at selected intermediate layers, preserving historical context without expanding the input token sequence. (b) In K-to-K Retrieval with \FGTB{} (see Sec. \ref{sec:kk_retrieval}\& \ref{sec:fgtb}), the module utilizes current keys to retrieve relevant historical features via address-space matching, applying a fixed Frame-Gap Temporal Bias (\FGTB{}) to down-weight stale history and minimize interference. (c) Finally, via Norm-Preserving Residual Loading (see Sec. \ref{sec:injection}), the retrieved history is injected into the current state through a rescaled residual update, enabling the frozen backbone to generate temporally consistent actions without parameter updates.}
  \label{fig:method}
\end{figure*}

\section{Related Works}

\subsection{Vision-Language-Action Models}
Vision--Language--Action (VLA) models map visual observations and language instructions to robot actions by leveraging pretrained vision--language backbones and large-scale robot demonstrations.
A key differentiator is the action generation paradigm: autoregressive decoders tokenize control and predict actions sequentially, as in RT-2 and OpenVLA~\cite{zitkovich2023rt2,kim2024openvla}; diffusion- or flow-style policies generate continuous trajectories or chunks to better capture multi-modality, as in $\pi_0$, CogACT, and DexVLA~\cite{black2024pi0,li2024cogact,wen2025dexvla}.
Efficient adaptation via fine-tuning and parameter-efficient updates further improves transfer to new tasks and robots~\cite{kim2025finetuningvla}.

Despite progress, many VLA inference pipelines remain largely \emph{memoryless} and do not explicitly retrieve long-horizon evidence, which is brittle in non-Markovian manipulation.

\subsection{Temporal Modeling and Inference in Robotics}
To mitigate temporal myopia, one line of work expands temporal inputs via frame stacking or video-style encoders; Octo and GR-2 follow this direction~\cite{octo2024,cheang2024gr2}. These approaches typically increase token length and inference latency, and they often require retraining with temporally structured inputs, thereby limiting plug-and-play adaptation for strong single-frame backbones.
A second line introduces explicit temporal interfaces: ContextVLA amortizes multi-frame context into compact representations~\cite{Jang2025ContextVLA}. MemoryVLA learns retrieval, fusion, and consolidation over a perceptual--cognitive memory bank~\cite{shi2025memoryvla}. HAMLET augments pretrained VLAs with history-aware tokens and a lightweight memory module, but still relies on fine-tuning~\cite{koo2025hamlet}.
Beyond these temporal interfaces, orthogonal approaches improve long-horizon coherence via motion or foresight, e.g., HiF-VLA~\cite{lin2025hifvla}.
Overall, prior temporal VLA solutions either pay for history with longer contexts and higher inference cost, or introduce additional temporal modules that often require extra training.
We therefore focus on TempoFit, a training-free, state-level retrofit that caches and reuses internal prefix $K/V$ across timesteps to inject history without longer input sequences or additional trainable components.

\section{Method}

\subsection{Preliminaries and Policy-Agnostic Setting}
\label{sec:prelim}

Modern Vision--Language--Action (VLA) policies typically function as a composite system.
Formally, we decompose a policy $\pi_\theta$ into two functional stages:
a pretrained vision--language backbone $\mathcal{F}_\phi$ and a downstream action head $\pi_\psi$, such that $\theta = \{\phi, \psi\}$.
At timestep $t$, the backbone encodes the current visual observation $o_t$ and language instruction $x$ into a sequence of latent representations $H_t$:
\begin{equation}
H_t = \mathcal{F}_\phi(o_t, x).
\label{eq:vla_backbone}
\end{equation}
Subsequently, the action head maps these representations to a predicted action chunk of horizon $n$:
\begin{equation}
\hat{a}_{t:t+n} \sim \pi_\psi(a_{t:t+n} \mid H_t).
\label{eq:vla_head}
\end{equation}
In standard inference, this process is \emph{Markovian}: $\mathcal{F}_\phi$ processes $o_t$ in isolation, resetting its internal state at every step.
The action head $\pi_\psi$ is architecture-agnostic and relies entirely on $H_t$ to capture the state.

\textbf{Our Focus.} In this work, we specifically target the \textbf{prefix encoding} phase within the backbone $\mathcal{F}_\phi$.
Rather than retraining $\pi_\psi$ or explicitly concatenating history to $o_t$ , which alters the input structure, TempoFit intervenes directly in the internal attention mechanism of $\mathcal{F}_\phi$.
By modifying the cached prefix $K/V$ states, we convert the memoryless mapping in Eq.~\eqref{eq:vla_backbone} into a history-aware encoding $\tilde{H}_t = \mathcal{F}_\phi(o_t, x, \mathcal{H}_{<t})$ that the frozen action head $\pi_\psi$ can consume transparently.

\subsection{Overview}
\label{sec:overview}

To address the temporal myopia of the frozen backbone $\mathcal{F}_\phi$ defined in Section \ref{sec:prelim}, our objective is to retrofit the policy with long-horizon consistency without fine-tuning parameters or expanding the input token context.
We propose \textbf{TempoFit}, a training-free framework that leverages internal prefix keys and values ($K/V$) as a model-native, layer-wise memory to inject historical context directly into the inference stream.
The overall pipeline of the proposed mechanism is illustrated in Figure \ref{fig:method}.
We first detail the construction of our layer-wise FIFO memory cache in Section \ref{sec:write}.
Next, Section \ref{sec:kk_retrieval} describes our parameter-free K-to-K retrieval mechanism, which is augmented by the Frame-Gap Temporal Bias (\FGTB{}) introduced in Section \ref{sec:fgtb} to suppress stale context.
Finally, Section \ref{sec:injection} explains how retrieved history is fused into the current state via norm-preserving residual loading to maintain inference stability under frozen weights.

\subsection{Memory Write: Layer-Wise FIFO KV Cache}
\label{sec:write}

To preserve historical evidence without expanding the input context length or introducing training,
we maintain a compact inference-time state directly in KV space.
Beyond \emph{what} to store, a practical question is \emph{where} to store it:
enabling temporal retrofitting at arbitrary depths can introduce history--present interference
and lead to large performance drops (Table.~\ref{tab:ablation_detailed}).

Building on evidence that Transformer representations are organized hierarchically across depth, where intermediate layers capture compositionally rich and transferable features and deeper layers become more specialized to the pretraining objective~\cite{tenney2019bertrediscoversclassicalnlp,meng2023locatingeditingfactualassociations}, we activate memory only in a small subset of intermediate layers. This design preserves transferability while reducing interference from task-specific representations.
Formally, consider an $L$-layer Transformer backbone and a memory-enabled layer subset
$\mathcal{L}_{\mathrm{mem}}\subset\{1,\dots,L\}$.
For each $l\in\mathcal{L}_{\mathrm{mem}}$, we maintain a FIFO buffer of capacity $C$ :
\begin{equation}
\mathcal{M}_l^{(t)}=\{(K_l^{(\tau)},V_l^{(\tau)},\tau)\}_{\tau\in\mathcal{T}_l^{(t)}} ,
\end{equation}
where $\mathcal{T}_l^{(t)}$ denotes the timesteps stored in the buffer.
Each entry corresponds to one past timestep $\tau$ and stores the \emph{prefix-time} projections
$(K_l^{(\tau)},V_l^{(\tau)})\in\mathbb{R}^{B\times H\times S\times d}$ produced during prefix encoding, where
$B$ is batch size, $H$ is the number of heads, $S$ is the number of prefix tokens, and $d$ is the per-head dimension.
We cache the tensors after linear projection and before applying rotary positional embeddings(RoPE).
Each timestep contributes $S$ prefix tokens.
At timestep $t$, once $(K_l^{(t)},V_l^{(t)})$ is computed, we append $(K_l^{(t)},V_l^{(t)},t)$ to $\mathcal{M}_l^{(t)}$
and evict the oldest entry if $|\mathcal{M}_l^{(t)}|>C$.
We cache only prefix-time $K/V$ for prefix tokens, excluding any action (suffix) tokens, and do not append any additional tokens to the input sequence.

\begin{table*}[t]
\vspace{2mm}
\centering
\caption{Ablation study of our temporal memory module on the \textsc{LIBERO-Long} benchmark with two backbone VLAs ($\pi_{0.5}$ and QwenGr00t). We report the per-task and average success rate (\%) across 10 tasks (50 trials each). ``Memory'' indicates whether temporal memory is enabled. ``Memory Type'' indicates whether it is not used (-), training-based ($\times$), or training-free ($\checkmark$). \textbf{Bold} indicates the best performance within each backbone group.}
\label{tab:libero_long_ablation}
\resizebox{\textwidth}{!}{%
\begin{tabular}{lccccccccccccc}
\toprule
\textbf{Method} & \textbf{Memory?} & \textbf{Training-free?} & \textbf{Avg. SR} & \makecell{Put soup\\and box\\in basket} & \makecell{Put box\\and butter\\in basket} & \makecell{Turn on\\stove and\\put pot} & \makecell{Put bowl\\in drawer\\and close} & \makecell{Put mugs\\on left\\and right\\plates} & \makecell{Pick book\\and place\\it in back} & \makecell{Put mug\\on plate,\\pudding\\right} & \makecell{Put soup\\and sauce\\in basket} & \makecell{Put both\\pots on\\stove} & \makecell{Put mug\\in\\microwave\\and close} \\
\midrule
Seer (scratch)~\cite{tian2025seer} & \textcolor{red}{$\times$} & - & 78.7 & 80.0 & 90.0 & 91.7 & 81.7 & 85.0 & 65.0 & 86.7 & 88.3 & 51.7 & 66.7 \\
Seer~\cite{tian2025seer}           & \textcolor{red}{$\times$} & - & 87.7 & 91.7 & 90.0 & 98.3 & \textbf{100} & 91.7 & 93.3 & 85.0 & 88.3 & 61.7 & 71.7 \\
UniVLA~\cite{bu2025univla}         & \textcolor{red}{$\times$} & - & 90.0 & \textbf{100} & 92.0 & 94.0 & 98.0 & 86.0 & \textbf{100} & 80.0 & \textbf{100} & 70.0 & 82.0 \\
OpenVLA-OFT~\cite{kim2025finetuningvla} & \textcolor{red}{$\times$} & - & 94.0 & 90.0 & \textbf{98.0} & 98.0 & 98.0 & 96.0 & \textbf{100} & 92.0 & \textbf{100} & 72.0 & 96.0 \\
MemoryVLA~\cite{shi2025memoryvla} & \textcolor{blue}{$\checkmark$} & \textcolor{red}{$\times$} & 93.4 & 92.0 & 96.0 & 96.0 & \textbf{100} & \textbf{100} & \textbf{100} & \textbf{96.0} & 96.0 & 62.0 & \textbf{96.0} \\
HiF-VLA \cite{lin2025hifvla} & \textcolor{blue}{$\checkmark$} & \textcolor{red}{$\times$} & 96.4 & 88.0 & \textbf{98.0} & \textbf{100} & \textbf{100} & \textbf{100} & \textbf{100} & \textbf{96.0} & \textbf{100} & \textbf{82.0} & \textbf{100} \\
\midrule
QwenGr00t \cite{starvla2025} & \textcolor{red}{$\times$} & - & 90.8 & 88 & 94.0 & \textbf{100} & 98.0 & 96.0 & \textbf{100} & 68.0 & 100.0 & 66.0 & \textbf{98.0} \\
\rowcolor{gray!15}
\textbf{TempoFit$_{\text{QwenGr00t}}$ (Ours)} & \textcolor{blue}{$\checkmark$} & \textcolor{blue}{$\checkmark$} & \textbf{94.4} & \textbf{100} & \textbf{98.0} & \textbf{100} & \textbf{100} & \textbf{100} & 98.0 & \textbf{80.0} & \textbf{92.0} & \textbf{88.0} & 88.0 \\
\midrule
$\pi_{0.5}$\cite{pi05_2025}  & \textcolor{red}{$\times$} & - & 92.6 & \textbf{100} & 96.0 & \textbf{98.0} & 96.0 & 96.0 & \textbf{100} & \textbf{96.0} & 90.0 & 58.0 & \textbf{96.0} \\
\rowcolor{gray!15}
\textbf{TempoFit$_{\pi_{0.5}}$ \ \ \ \ \ \ \ (Ours)} & \textcolor{blue}{$\checkmark$} & \textcolor{blue}{$\checkmark$} & \textbf{96.6} & \textbf{100} & \textbf{100} & \textbf{98.0} & \textbf{98.0} & \textbf{100} & \textbf{100} & \textbf{96.0} & \textbf{96.0} & \textbf{84.0} & \textbf{96.0} \\
\bottomrule
\end{tabular}%
}
\end{table*}

\subsection{K-to-K Retrieval: Memory Update via Address-Space Matching}
\label{sec:kk_retrieval}

In scaled dot-product attention, an output is produced by matching a query against a set of keys
and reading a weighted sum of the corresponding values, i.e., keys/values form a model-native, content-addressable memory table~\cite{vaswani2017attention}.
This view is made explicit in key--value memory networks, where keys serve as \emph{addresses} and values store \emph{content}~\cite{miller2016keyvalue}.
Therefore, if we cache prefix-time per-layer $(K,V)$ tensors across timesteps, the most training-free and interface-consistent way to retrieve historical evidence
is to match \emph{within the same key space} in which the pretrained Transformer already performs addressing.
This is also aligned with interpreting attention as associative retrieval, which can be cast as a modern Hopfield-style update where retrieval is determined by similarity in the stored-pattern space~\cite{ramsauer2021hopfield}.

For a memory-enabled layer $l\in\mathcal{L}_{\mathrm{mem}}$, we concatenate historical prefix keys/values as $(K_l^{\mathrm{hist}},V_l^{\mathrm{hist}})$.
We then treat the \emph{current} prefix keys $K_l^{(t)}$ as retrieval queries and compute per-head logits
by key-to-key similarity:
\begin{equation}
A_{l,h}^{\mathrm{kk}}=\frac{K_{l,h}^{(t)}(K_{l,h}^{\mathrm{hist}})^\top}{\sqrt{d}} + \mathrm{Mask},
\label{eq:kk_logits}
\end{equation}
followed by $W_{l}=\mathrm{Softmax}(A_l^{\mathrm{kk}})$ and context readout
$K_l^{\mathrm{ctx}}=W_lK_l^{\mathrm{hist}}$, $V_l^{\mathrm{ctx}}=W_lV_l^{\mathrm{hist}}$.
We operate on \emph{pre-RoPE} projections so that matching is primarily content-driven; positional encoding (RoPE) is subsequently applied using the \emph{current} positions~\cite{su2024roformer}.

In a training-free setting, we avoid introducing new learned query projections or gates.
Using $K^{(t)}$ to query $K^{\mathrm{hist}}$ performs \emph{address-space matching} under the same projection $W_K$ as the frozen backbone,
yielding retrieval that is compatible with the pretrained attention geometry and does not depend on the action head or denoising-step-specific queries.
Empirically, this reduces cross-stage interference while preserving plug-and-play deployment.

K-to-K retrieval is parameter-free, layer-local, and model-native: it reuses the Transformer’s existing addressing metric
and can be interpreted as an associative memory read in the key space~\cite{vaswani2017attention,ramsauer2021hopfield}.
It complements the subsequent standard self-attention by producing temporally enriched $(K,V)$ states
that downstream action heads can consume unchanged.

\subsection{Time-Biased Retrieval: Frame-Gap Temporal Bias\ (\FGTB{})}
\label{sec:fgtb}


However, naively retrieving over the entire cache can over-emphasize stale cues and induce history--present interference.
Prior long-context and recurrent-memory Transformers often mitigate this issue by explicitly managing old memories, such as compressing distant context and expiring stale states~\cite{rae2020compressive,sukhbaatar2021expire}.
Since TempoFit is training-free and cannot learn an explicit gating policy, we instead introduce a fixed and interpretable recency prior to down-weight outdated history.

Inspired by positional bias formulations in natural language processing ~\cite{press2022alibi}, we propose \FGTB{}, a \emph{frame-gap} temporal bias added to the K-to-K retrieval logits.
Unlike token-distance biases defined over text positions, \FGTB{} is defined over \emph{timestep gaps} and serves as a lightweight,
training-free safeguard that keeps decisions present-dominant.

Concretely, we augment the K-to-K retrieval logits in Eq.~\eqref{eq:kk_logits} with an additive linear bias:
\begin{equation}
\mathrm{Bias}_{l,h}(t,\tau)=-\beta \cdot m_h \cdot |t-\tau| \cdot \alpha_S,
\label{eq:fgtb}
\end{equation}
where $m_h$ follows a head-wise slope schedule inspired by ALiBi~\cite{press2022alibi},
$\beta$ controls the decay strength, and $\alpha_S$ maps frame gaps to token scale (default $\alpha_S=S$).
We then use $A_{l,h}=A_{l,h}^{\mathrm{kk}}+\mathrm{Bias}_{l,h}(t,\tau)$, yielding a simple ``content + recency prior'' retrieval rule that keeps decisions present-dominant.


\FGTB{} reduces interference from stale history while retaining soft access to earlier-but-relevant evidence.
Its effect is interpretable: the bias enforces a monotonic decay with the frame gap $|t-\tau|$ (Eq.~\ref{eq:fgtb}) and is tunable via $\beta$, making it well-suited for training-free temporal retrofitting.

\subsection{KV Injection: Norm-Preserving Residual Loading}
\label{sec:injection}

After retrieval (Sec.~\ref{sec:kk_retrieval}) with \FGTB{} (Sec.~\ref{sec:fgtb}), we obtain
$(K_l^{\mathrm{ctx}}, V_l^{\mathrm{ctx}})$ in the \emph{same} key/value space as the frozen backbone.
The injection mechanism must therefore (i) expose this context through standard self-attention,
(ii) introduce no trainable parameters while keeping tokenization, tensor shapes, and masks unchanged,
and (iii) keep the resulting prefix cache reusable by arbitrary action heads.
A straightforward alternative is to append retrieved features as extra ``virtual tokens''~\cite{li2021prefixtuning};
however, concatenation changes the attended length and softmax normalization, and incurs additional
compute/memory that scales with the added context---a mismatch for training-free retrofitting under frozen weights.

We instead inject history by \emph{updating the existing KV table} via residual loading:
\begin{equation}
\tilde K_l^{(t)} \;=\; K_l^{(t)} + K_l^{\mathrm{ctx}},\qquad
\tilde V_l^{(t)} \;=\; V_l^{(t)} + V_l^{\mathrm{ctx}}.
\label{eq:kv_residual_loading}
\end{equation}
This operation is parameter-free and preserves all shapes and masks, so subsequent attention consumes history
with the original computation.
However, the additive update can shift KV magnitudes away from the distribution expected by downstream frozen
layers, potentially destabilizing softmax. To mitigate this distribution shift, we apply a \textbf{norm-preserving}
rescaling that projects the fused tensor back to the original per-token $\ell_2$ norm:
\begin{equation}
\tilde K_l^{(t)} \;\leftarrow\;
  \tilde K_l^{(t)} \cdot
  \frac{\lVert K_l^{(t)}\rVert}
       {\max\!\bigl(\lVert\tilde K_l^{(t)}\rVert,\,\epsilon\bigr)},
\label{eq:norm_preserve}
\end{equation}
and analogously for $\tilde V_l^{(t)}$.
This constrains the injection to a \emph{directional} update---history can steer the effective KV associations
without inflating/deflating scale---and adds negligible overhead (two norms and one element-wise multiply).
Overall, the procedure is consistent with inference-time, non-parametric memory augmentation (retrieval without
weight updates)~\cite{grave2017cache,khandelwal2020knnlm,wu2022memorizing}, while remaining KV-native
and layer-local: we modify the current-step KV table rather than introducing extra tokens or a separate fusion head.

\section{EXPERIMENTS}

In this section, we design experiments to address the following research questions (RQs):
\begin{itemize}

  \item \textbf{RQ1:} How does \textbf{TempoFit} perform compared to SOTA methods on challenging long-horizon benchmarks like \textsc{LIBERO-Long} and \textsc{CALVIN}?

  \item \textbf{RQ2:} Can \textbf{TempoFit} reduce the redundancy and inefficiency of conventional observation-level history while remaining scalable to longer temporal horizons?

  \item \textbf{RQ3:} How do different components of \textbf{TempoFit}, such as selected layers and \FGTB{}, contribute to its overall performance?

  \item \textbf{RQ4:} Can \textbf{TempoFit} handle long-horizon tasks on real-world robotic platforms effectively?
\end{itemize}

\begin{table}[t]
\centering
\caption{Performance comparison on the CALVIN D-D and CALVIN ABC-D benchmarks.
We report the average number of successfully completed tasks across five consecutive instructions.
\textbf{Bold} indicates the best performance within each benchmark.}
\label{tab:calvin}
\vspace{-1mm}
\setlength{\tabcolsep}{3.2pt}
\renewcommand{\arraystretch}{1.05}
\scriptsize
\begin{tabular}{lcccccc}
\toprule
\textbf{Method} & 1 & 2 & 3 & 4 & 5 & \textbf{Avg. Len. $\uparrow$} \\
\midrule
\multicolumn{7}{l}{\textit{CALVIN D-D}} \\
\midrule
$\pi_0$~\cite{black2024pi0} & 84.8 & 70.4 & 55.9 & 46.6 & 37.7 & 2.95 \\
QwenPI~\cite{starvla2025} & 90.9 & 79.5 & 69.6 & 62.2 & 55.4 & 3.58 \\
\textbf{QwenGR00T}~\cite{starvla2025} & \textbf{92.5} & \textbf{83.9} & 74.4 & 67.9 & 59.8 & 3.78 \\
\rowcolor{gray!15}
\textbf{TempoFit$_{\text{QwenGr00t}}$} (Ours) & 92.0 & 83.8 & \textbf{75.7} & \textbf{70.3} & \textbf{62.3} & \textbf{3.84} \\
\midrule
\multicolumn{7}{l}{\textit{CALVIN ABC-D}} \\
\midrule
\textbf{$\pi_{0.5}$}\cite{pi05_2025} & \textbf{93.2} & 84.6 & 76.7 & 68.8 & 61.4 & 3.83 \\
\rowcolor{gray!15}
\textbf{TempoFit$_{\pi_{0.5}}$} (Ours) & 93.0 & \textbf{84.8} & \textbf{77.3} & \textbf{69.4} & \textbf{62.0} & \textbf{3.87} \\
\bottomrule
\end{tabular}
\vspace{-2mm}
\end{table}

\subsection{Overall Performance}

\textbf{Experimental Setups.} We evaluate our method on two long-horizon benchmarks: \textsc{LIBERO-Long} \cite{liu2023libero} and \textsc{CALVIN} \cite{mees2022calvin}. \textsc{LIBERO-Long} comprises ten multi-subgoal manipulation tasks across diverse scenes. For \textsc{CALVIN}, we conduct evaluations under two distinct settings to comprehensively assess the model: 1) the in-domain $D \rightarrow D$ setting, where policies are trained on demonstrations from environment D and evaluated on held-out sequences in the same environment D, measuring consecutive multi-task performance without cross-environment generalization; and 2) the cross-domain ABC-D setting, where policies are trained on environments A-C and evaluated on the unseen environment D to assess generalization on consecutive tasks. All experiments are conducted under a multi-view setup using both the primary and wrist cameras. For our baselines, we directly adopt the Qwen-GR00T \cite{starvla2025} and Open$\pi$ \cite{black2024pi0} checkpoints for the $D \rightarrow D$ evaluation, and the $\pi_{0.5}$ checkpoint from the RLinf project \cite{yu2025rlinfflexibleefficientlargescale, pi05_2025} for the ABC-D evaluation.

\textbf{Implementation Details.} We implement our framework following the standard fine-tuning setup for VLA models. For the baseline comparisons, we prioritize reproducibility and fairness by directly utilizing the official checkpoints provided by the respective state-of-the-art projects without heuristic hyperparameter tuning. Specifically, we evaluate the $\pi_{0.5}$ checkpoint from the RLinf project \cite{yu2025rlinfflexibleefficientlargescale, pi05_2025} and the QwenGR00T checkpoint from the StarVLA project \cite{starvla2025} on the \textsc{LIBERO} and \textsc{CALVIN} benchmarks. Regarding temporal modeling, we fix the history capacity to 8 frames across all our experiments to ensure the model captures sufficient temporal context. All evaluations are performed under the consistent multi-view setup described in the experimental settings.

\textbf{Result Analysis.} \textbf{1) \textsc{LIBERO-Long}:} As shown in Table \ref{tab:libero_long_ablation}, we present the detailed performance of \textbf{TempoFit} across 10 tasks in the \textsc{LIBERO-Long} benchmark. We evaluate our method across two strong memoryless baselines ($\pi_{0.5}$ and QwenGR00T) over 500 trials. Compared to the vanilla baselines, our approach achieves substantial gains: applying \textbf{TempoFit} to the $\pi_{0.5}$ backbone increases the average success rate from 92.6\% to 96.6\%, representing a 4.0\% absolute improvement. Similarly, for QwenGR00T, it yields a 3.6\% absolute improvement (from 90.8\% to 94.4\%). Crucially, this highlights a fundamental advantage of our \emph{training-free} paradigm: by strictly preserving the original pretrained weights, \textbf{TempoFit} can directly harness the potent visual-linguistic representations and generalization capabilities of cutting-edge backbones like $\pi_{0.5}$. Consequently, it enables these powerful single-frame models to surpass representative \emph{training-based} temporal approaches such as MemoryVLA (93.4\%) and even edge out HiF-VLA (96.4\%). The performance boost is especially pronounced in challenging subgoals that demand strict cross-stage temporal association (e.g., success on ``Put both pots on stove'' surges from 58.0\% to 84.0\% for $\pi_{0.5}$). This underscores that seamlessly retrofitting existing strong VLAs with our layer-wise memory is a highly effective route to robust temporal reasoning, fully unlocking their pretrained potential without the computational overhead or catastrophic forgetting risks associated with model retraining.

\textbf{2) \textsc{CALVIN}:} We further evaluate \textbf{TempoFit} on \textsc{CALVIN} under two complementary settings (Table~\ref{tab:calvin}): the in-domain \textit{D-D} setting and the cross-domain \textit{ABC-D} setting (generalization to the unseen environment D).
In \textit{D$\rightarrow$D}, applying TempoFit to the strong QwenGR00T checkpoint improves the average task length from 3.78 to 3.84. Notably, the gain concentrates on later instructions where temporal credit assignment and partial observability become dominant: while the first two instructions remain essentially unchanged, TempoFit yields consistent improvements on instructions 3--5 , indicating better long-horizon retention rather than short-horizon action selection.
In \textit{ABC$\rightarrow$D}, TempoFit also improves the RLinf $\pi_{0.5}$ checkpoint from 3.83 to 3.87, with small but consistent gains on later instructions.
Overall, these results suggest that \emph{state-level} temporal retrofitting can translate into more reliable multi-step execution as the horizon grows: by reusing cached intermediate-layer prefix $K/V$ and enforcing a fixed recency prior with \FGTB{}, TempoFit better disambiguates temporally aliased states and mitigates cross-stage fragmentation, while preserving the original single-frame inference graph and requiring no additional training.


\subsection{Inference Efficiency and Horizon Scalability}
Inference-time scalability is critical for closed-loop manipulation, where control must remain near real time as temporal context grows. We therefore measure per-step latency and peak GPU memory while increasing the KV-cache capacity $C$ on \textsc{LIBERO-Long} (Table~\ref{tab:efficiency}). TempoFit adds only a small overhead over the 1-frame baseline (71.2\,ms): 73.4\,ms at $C{=}4$ and 74.4\,ms at $C{=}8$, and consistently remains 86.8\,ms even at $C{=}32$ with $\le$1.10$\times$ memory. This mild scaling stems from caching only prefix $K/V$ at selected layers. In contrast, naive frame stacking grows rapidly (94.8\,ms/3.54$\times$ at 4 frames; 176.3\,ms/7.19$\times$ at 8 frames).

\begin{table}[t]
\vspace{2mm}
\centering
\caption{\textbf{Efficiency analysis.} Average latency and peak memory usage measured on \textsc{LIBERO-long}
datasets. Both metrics are computed at each timestep within an episode and then averaged.
All measurements were on an NVIDIA RTX5090 GPU. $\downarrow$ indicates lower values are better.}
\label{tab:efficiency}
\setlength{\tabcolsep}{3.6pt}
\renewcommand{\arraystretch}{0.95}
\scriptsize
\begin{tabular}{lccc}
\toprule
Method & History Length & Latency (ms, $\downarrow$) & Peak memory (MB, $\downarrow$) \\
\midrule
$\pi_{0.5}$ \cite{pi05_2025} & 1  & 71.2 ($1.00\times$) & 6{,}396 ($1.00\times$) \\
\midrule
+  Multi-frames         & 4  & 94.8 ($1.33\times$) & 22{,}640 ($3.54\times$) \\
\rowcolor{gray!15}
\textbf{+ TempoFit (Ours)}          & 4  & \textbf{73.4} ($\mathbf{1.02\times}$) & \textbf{6{,}498} ($\mathbf{1.02\times}$) \\
\midrule
+  Multi-frames         & 8  & 176.3 ($2.48\times$) & 45{,}980 ($7.19\times$) \\
\rowcolor{gray!15}
\textbf{+ TempoFit (Ours)  }        & 8  & \textbf{74.4} ($\mathbf{1.04\times}$)& \textbf{6{,}600} ($\mathbf{1.03\times}$) \\
\rowcolor{gray!15}
\textbf{+ TempoFit (Ours)  }        & 16 & \textbf{81.4} ($\mathbf{1.13\times}$)& \textbf{6{,}761} ($\mathbf{1.06\times}$) \\
\rowcolor{gray!15}
\textbf{+ TempoFit (Ours)  }        & 32 & \textbf{86.8} ($\mathbf{1.21\times}$) & \textbf{7{,}030} ($\mathbf{1.10\times}$) \\
\bottomrule
\end{tabular}
\end{table}

\begin{table}[t]
\centering
\caption{\textbf{Detailed ablation study} of TempoFit on \textsc{LIBERO-Long} ($\pi_{0.5}$ backbone). Each row modifies one design choice from the full method (last row). We report average success rate (\%) over 500 trials.}
\label{tab:ablation_detailed}
\vspace{-1mm}
\setlength{\tabcolsep}{5pt}
\renewcommand{\arraystretch}{0.95}
\scriptsize
\begin{tabular}{lc}
\toprule
\textbf{Configuration} & \textbf{Avg. SR (\%)} \\
\midrule
Baseline (no memory)                        & 92.6 \\
\midrule
\multicolumn{2}{l}{\textit{Component contribution}} \\
\quad + KV Memory only                      & 93.8 \\
\quad + KV Memory + \FGTB{}                 & \textbf{96.6} \\
\midrule
\multicolumn{2}{l}{\textit{Retrieval strategy}} \\
\quad Q-to-K retrieval                      & 93.3 \\
\quad K-to-K retrieval (Ours)               & \textbf{96.6} \\
\midrule
\multicolumn{2}{l}{\textit{Injection strategy}} \\
\quad Concatenation                         & 0.8 \\
\quad Residual loading \textbf{w/o} norm-preserving         & 90.2 \\
\quad Residual loading \textbf{w/} norm-preserving (Ours)   & \textbf{96.6} \\
\midrule
\multicolumn{2}{l}{\textit{Layer selection}} \\
\quad All layers (0-17)                           & 74.2 \\
\quad Bottom layers only (9-17)                   & 59.8 \\
\quad Top layers only (0-8)                       & 89.4 \\
\quad Intermediate layers (Ours)                  & \textbf{96.6} \\
\midrule
\multicolumn{2}{l}{\textit{History capacity $C$}} \\
\quad $C=4$                                 & 95.2 \\
\quad $C=8$ (Ours)                          & \textbf{96.6} \\
\quad $C=16$                                & 96.2 \\
\quad $C=32$                                & 95.2 \\
\bottomrule
\end{tabular}
\vspace{-2mm}
\end{table}

\begin{figure*}[t]
\vspace{2mm}
  \centering
      \includegraphics[width=\textwidth]{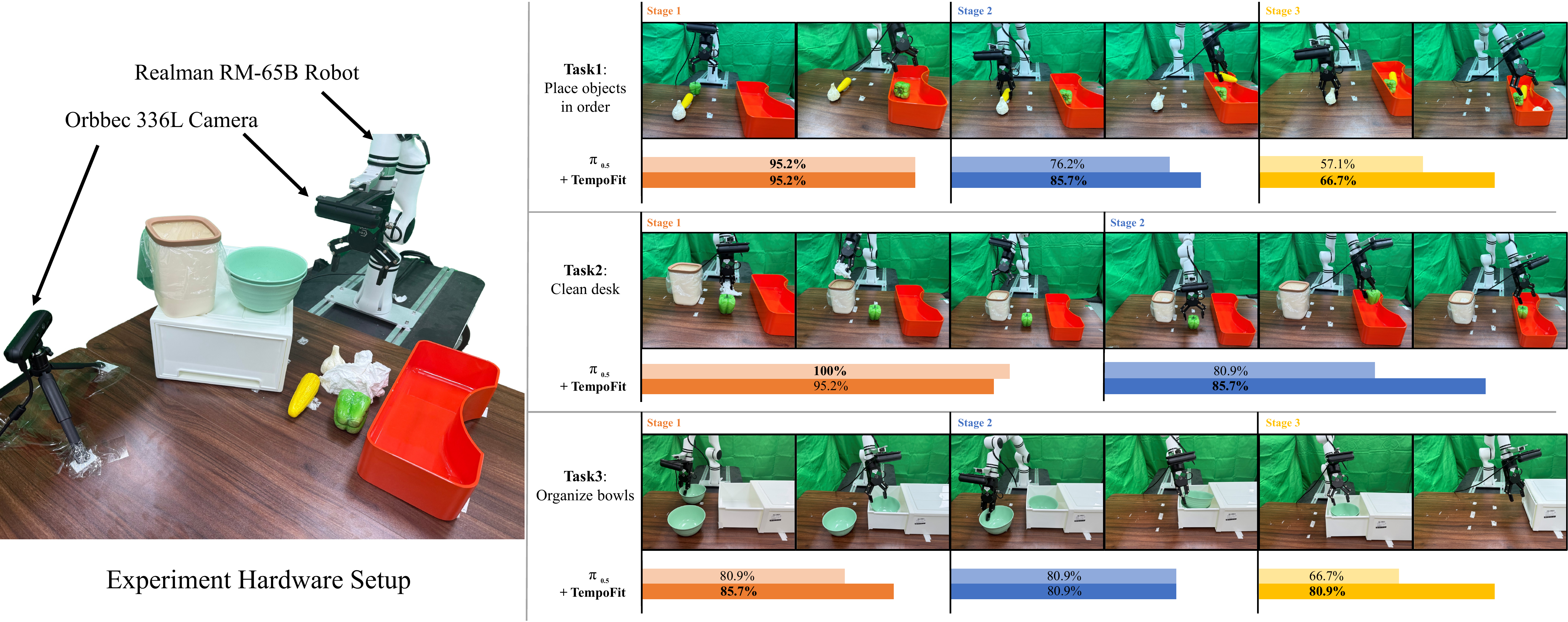}
  \caption{\textbf{Real-world evaluation on Realman RM-65B.} \textbf{Left.} Hardware and multi-view sensing setup. \textbf{Right.} Quantitative success rates and/or qualitative rollouts on three long-horizon manipulation tasks.}
  \label{fig:real_robot}
  \vspace{-2mm}
\end{figure*}

\subsection{Ablation Studies}

We conduct a detailed ablation of \textbf{TempoFit} on \textsc{LIBERO-Long} with a $\pi_{0.5}$ backbone (Table~\ref{tab:ablation_detailed}; 500 trials). Each row modifies a single design choice from the full model (last row), allowing us to isolate the contribution of core components and key implementation choices.

\textbf{Component contribution.}
Starting from the memoryless baseline (92.6\%), enabling \emph{KV memory} alone yields a modest improvement to 93.8\% (+1.2), indicating that reusing prefix-time internal states already provides useful temporal evidence.
However, this gain is amplified only when we introduce \FGTB{}: adding \FGTB{} on top of KV memory boosts performance to 96.6\% (+2.8 over KV-only), supporting the claim that an explicit, fixed recency bias is critical for training-free temporal retrofitting to keep decisions present-dominant and suppress stale-history interference.

\textbf{Retrieval strategy.}
We further ablate the addressing mechanism used for memory readout. Replacing our KV-native \emph{K-to-K} retrieval with a \emph{Q-to-K} alternative reduces success to 93.3\%, whereas K-to-K restores the full performance (96.6\%). This suggests that matching in the backbone’s native key-address space is substantially more compatible under frozen weights.

\textbf{Injection strategy.}
How retrieved history is injected is also decisive. Naively injecting history via \emph{concatenation} collapses performance (0.8\%), implying that expanding the effective attended context without retraining can severely miscalibrate the frozen attention computation. Residual loading is markedly more stable, but removing the proposed \emph{norm-preserving} rescaling still causes a clear drop (96.6\% $\rightarrow$ 90.2\%). With norm-preserving enabled, residual loading achieves the best result (96.6\%), consistent with the motivation that controlling the KV magnitude is essential to avoid distribution shift introduced by additive updates in a training-free setting.

\textbf{Layer selection and history capacity.}
Finally, we examine where to enable memory and how much history to store. Activating memory at all layers (0--17) substantially degrades performance (74.2\%), and restricting memory to only a partial depth range is also suboptimal (0--8: 89.4\%; 9--17: 59.8\%). In contrast, enabling memory only at the selected \emph{intermediate} layers achieves the best performance (96.6\%), supporting our layer-selective design to balance temporal continuity with minimal interference to present-step control. Varying capacity reveals that a moderate history length works best: $C=8$ achieves 96.6\%, while both smaller and larger capacities slightly underperform ($C=4$: 95.2\%; $C=16$: 96.2\%; $C=32$: 95.2\%), suggesting diminishing returns and increased redundancy/staleness when the cache grows too large.

\subsection{Real-World Robotic Platforms}
\textbf{Experiment setups.} To evaluate the effectiveness of our approach in real-world applications, we conduct real-world experiments using the Realman RM-65B robot. As shown in Fig. 5, a Orbbec 336L camera captures the scene from a third-view, while an additional USB camera is mounted  on the robot’s wrist for egocentric observations. We collect three long-horizon tasks, each with 100 demonstrations involving diverse manipulation primitives, including pick, place and push.

\textbf{Real-World Task Performance:} For real-world environments, we train the baseline $\pi_{0.5}$\cite{pi05_2025} for every task and evaluate performance by averaging success rates over 20 trials per task across three long-horizon manipulation tasks: sequentially placing three vegetables into a tray (Task 1), cleaning a desk by disposing of tissue and then placing a pepper into a box (Task 2), and putting both green bowls into a drawer and closing it (Task 3). These long-horizon tasks require the model to maintain action consistency across stages and correctly associate temporal states. As shown in Fig.\ref{fig:real_robot}, the baseline suffers clear degradation as the number of subtasks grows—for example, achieving only 57.1\% full-sequence success on Task 1 despite near-perfect first-subtask performance, and dropping to 66.7\% on the full Task 3 sequence, often stalling or repeating actions at later stages. This is likely due to state aliasing between visually similar objects (e.g., two identical green bowls) and subtle post-action visual changes that a memoryless policy fails to track. In contrast, \textbf{TempoFit} benefits from its layer-wise temporal KV retrieval with \FGTB{}, enabling reliable detection of completed subtask transitions and robust cross-stage execution, improving full-task success rates by +9.5\% on Task 1 (57.1\%→66.7\%), +4.8\% on Task 2 (81.0\%→85.7\%), and +14.3\% on Task 3 (66.7\%→81.0\%), yielding a +9.5\% average improvement across all real-world tasks. 

\section{CONCLUSIONS}

We introduced \textbf{TempoFit}, a training-free temporal retrofitting module that upgrades pretrained single-frame VLAs to be history-aware by reusing their internal attention state. The method caches prefix K/V at a small subset of intermediate layers, retrieves past evidence via K-to-K address-space matching with \FGTB{}, and injects the retrieved context through pre-attention residual loading, preserving the original tokenization, backbone parameters, and action head.
Across \textsc{LIBERO-Long}, \textsc{CALVIN}, and real-world Realman RM-65B tasks, \textbf{TempoFit} improves long-horizon coherence and success while adding only minor inference overhead compared to costly frame stacking.

\textbf{Limitations.} Our current implementation uses fixed choices for layer subset, cache capacity, and decay slope; performance can degrade if history becomes dominated by irrelevant frames or if the task requires very long-term planning beyond the cache horizon. Future work will explore adaptive memory selection, automatic layer discovery across backbones, and integrating KV-native temporal retrofitting with higher-level recovery or subgoal mechanisms.

\bibliographystyle{IEEEtran}
\bibliography{arxiv}

\end{document}